# Prediction of Temperature and Rainfall in Bangladesh using Long Short Term Memory Recurrent Neural Networks

Mohammad Mahmudur Rahman Khan
*Department of ECE*
Vanderbilt University
Nashville, United States
mohammad.mahmudur.rahman.khan@vanderbilt.edu

Md. Abu Bakr Siddique
*Department of EEE*
International University of Business Agriculture and Technology
Dhaka, Bangladesh
absiddique@iubat.edu

Shadman Sakib
*Department of EECS*
University of Hyogo
Himeji, Japan
sakibshadman15@gmail.com
shadman.sakib@ieee.org

Anas Aziz
*Department of AE*
Military Institute of Science and Technology
Dhaka, Bangladesh
anas.aziz@ae.mist.ac.bd

Ihtyaz Kader Tasawar
*Department of EEE*
BRAC University
Dhaka, Bangladesh
ihtyaztasawar@gmail.com
ihtyaztasawar@ieee.org

Ziad Hossain
*Department of CSE*
North South University
Dhaka, Bangladesh
ziadhossain98@gmail.com

*Abstract*—Temperature and rainfall have a significant impact on the economic growth as well as the outbreak of seasonal diseases in a region. In spite of that inadequate studies have been carried out for analyzing the weather pattern of Bangladesh implementing the artificial neural network. Therefore, in this study, we are implementing a Long Short-term Memory (LSTM) model to forecast the monthwise temperature and rainfall by analyzing 115 years (1901-2015) of weather data of Bangladesh. The LSTM model has showed a mean error of -0.38°C in case of predicting the monthwise temperature for 2 years and -17.64mm in case of predicting the rainfall. This prediction model can help to understand the weather pattern changes as well as studying seasonal diseases of Bangladesh whose outbreaks are dependent on regional temperature and/or rainfall.

## I. INTRODUCTION

A natural phenomenon such as the onset of heavy rainfall may create an increased possibility for disastrous events to occur. For instance, flooding might occur which is often the underlying reason for the accelerated spread of water-borne as well as vector-borne diseases. Moreover, floods can also lead to excessively-damaged crops, the destruction of people's properties, and put lives at risk. In addition to this, efficient road transportation and everyday activities could face incredible resistance by the long-duration heavy rainfall. In mid-2019, mainly the Northern area of Bangladesh had to face severe consequences due to the flood initiated by heavy rainfall. The flood had caused numerous people to suffer from cholera, diarrhea, dysentery, skin diseases, and many more water-borne diseases. Approximately 0.5 million hectares of croplands were reported to be destroyed. Also, about 0.6 million houses and 7000 kilometers of roadway were mostly damaged severely and the rest completely destroyed. The death count from the flood was 75.

Weather conditions such as the appearance of little to no rainfall in an area cause drought which in turn results in severe deterioration of crop yield. Additionally, extreme temperature conditions can also introduce various problems. Uneasiness and various health issues in human beings can result from very high or low temperatures which can eventually cause an increment in morbidity as well as the death rate. For instance, serious health issues such as heat stroke as well as diarrhea affected those working outdoors, aged people and children. This was the end result of the severe heatwave which hit the Southern part of Bangladesh in mid-2019. Also, at the end of 2019, cold temperatures caused around 50 people to lose their lives having diagnosed with respiratory infections and mostly diarrhea from the rotavirus. Moreover, the hospitals became over-crowded because people suffered from dehydration, pneumonia, and influenza. Besides, deviation from normal ranges of temperature can disrupt the optimum enzyme activity in plants and soil [1], [2] the moisture content of the soil, and the rate of soil respiration [3]. Therefore, a significant reduction of crop growth might be observed indicating a downfall in the agricultural sector.

Since the agriculture sector is a weighty contributing factor to a country's economy, both heavy rainfall and abnormal temperatures can have a detrimental effect on it. Moreover, the mortality rate might as well increase. Due to the flood in Bangladesh, crops worth BDT 11.52 billion were damaged in 2019. The need for avoiding such massive loss has compelled weather forecasting to become a necessity. Precautionary as well as preventive actions such as flood prevention methods, water resource management, etc. can be done if accurate weather prediction information is available.

The presence of non-linearity in temperature as well as rainfall patterns introduces difficulties in the accurate prediction of it. Previously, prediction approaches driven by theory were proposed. However, its complexity and need for great computing power did not allow the increased accuracy of these approaches. Conversely, proposed approaches that are driven by data are performing significantly better than the aforementioned method. Statistical methods along with machine-learning approaches mainly constitute the data-driven models. A good portion of weather prediction researches was initially done by statistical approaches. The additive variant of the Holt-Winters approach was applied by Manideep *et al.* [4] for rainfall prediction. ARIMA model was implemented for predicting the rainfall in Allahabad city by Graham *et al.* [5]. However, the important drawback of statistical approaches was their inapplicability in the prediction of the non-linear pattern-based phenomenon. Besides, the appropriateness of the machine learning methods to non-linear applications had made them the appropriate choice for the research work associated with weather prediction. ANNs, GEP, etc. are some of the machine-learning (ML) techniques which had been applied in rainfall estimation. Mehdizade *et al.* [6] proposed that ANNs have more preference compared to GEP in the area of rainfall estimation. Dash *et al.* [7] focused on the rainfall prediction in Kerala, India using the SLFN approach. An important consideration in the time-dependent phenomenon is that the previous state of estimation has to be taken into account which is a weakness of these approaches. To solve this issue, deep learning (DL) approaches such as RNN is used. Most recently, techniques associated with DL have been gaining large popularity because feature extraction from the data is executed automatically, with lower computational complexity and immense applicability. RNNs were used in predicting rainfall behavior due to its suitability in data possessing a temporal part. However, the problem regarding the vanishing of the gradient in long-term dependencies is the downside of RNNs. Therefore, this drawback was addressed by another method named LSTM. There are several fields where LSTM had been applied such as financial market prediction [8], processing of natural language [9], wind turbine fault diagnosis [10], predicting quality of air [11], etc.

A number of studies have been done in the prediction of rainfall or temperature applying LSTM. Caihong *et al.* [12] had simulated the rainfall-runoff process related to past flood occurrences in Northern China applying ANN as well as LSTM models. Their results indicated that the LSTM approach had more stability and displayed superior performance. Besides, by utilizing the temperature data recorded from the international airport in Beijing, the temperature at every half-hour was predicted by Li *et al.* [13] in their work where stacked LSTM approach was applied. Liao *et al.* [14] worked on an LSTM based model which could predict the rainfall 0-2 days earlier. The basin of the Jinjiang River in the Southeast area of China was taken as the study region. Moreover, both RNN, as well as the LSTM approach, was applied to estimate monthly rainfall [15]. The average of the monthly rainfall data of India was used as the dataset. Results displayed that LSTM performed significantly better than RNN in this study. Poornima *et al.* [16] proposed an RNN based intensified LSTM approach for rainfall prediction. 34 years of rainfall data of Hyderabad, India was utilized as the dataset used for training the model. Furthermore, using convolutional LSTM, Chen *et al.* [17] carried out the nowcasting of precipitation. In conclusion, this approach performed better than FC-LSTM as well as the ROVER algorithm. A study [18] which modeled rainfall-runoff by the LSTM approach used multiple catchments of the CAMELS dataset. The obtained accuracy was somewhat similar to the successful SAC-SMA model combined with the Snow-17 snow routine.

In this research, a Long Short-term Memory (LSTM) network is implemented as the prediction network to forecast the monthwise temperature and rainfall. The study explores 115 years (1901-2015) of the time series weather data of Bangladesh to provide accurate predictions and to reduce the mean error. Figure 1 illustrates the workflow of the proposed method. The organization of the rest of the paper is ordered according to this: Section 2 describes the detailed methodology for the prediction of temperature and rainfall comprising of a few subsections: data set details, an overview of the LSTM network, and the basic LSTM block. Experimental results are discussed in Section 3, and subsequently, Section 4 provides the conclusion and future works.

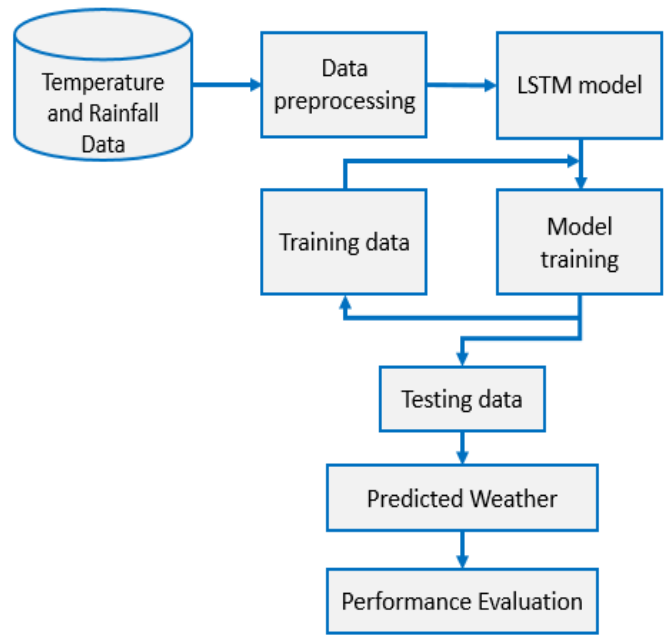

Fig. 1. Workflow of the proposed method

## II. METHODOLOGY

### A. Data

The monthly average temperature and rainfall data for 115 years of Bangladesh's weather are collected from Kaggle public domain dataset known as "Bangladesh Weather Dataset" [19]. This dataset contains monthly average temperature and rainfall data of Bangladesh from January 1901 to December 2015. For instance, the monthly average temperature and rainfall data for May 1901 are 27.8892-degree Celsius and 267.215 millimeters respectively. Again, the monthly average temperature and rainfall data for January 2014 are 17.1088-degree Celsius and 0.1202 millimeters respectively. Figure 2 shows the average month-wise temperature and rainfall data for all 115 years in Bangladesh respectively.

### B. Long Short-Term Memory (LSTM) Networks

Long Short-Term Memory neural network or more simply known as LSTM is considered as an enhanced category of Recurrent Neural Network (RNN). Hochreiter and Schmidhuber [20] were the pioneers of LSTM; responsible for its establishment in the year 1997. LSTM has experienced an increasing prominence over the years and has been used in a wide variety of work and problems. LSTM stands out from standard RNN particularly because of the architecture's ability to hold onto information for large periods of time as opposed to RNN. This ability is termed as a long-term dependency.

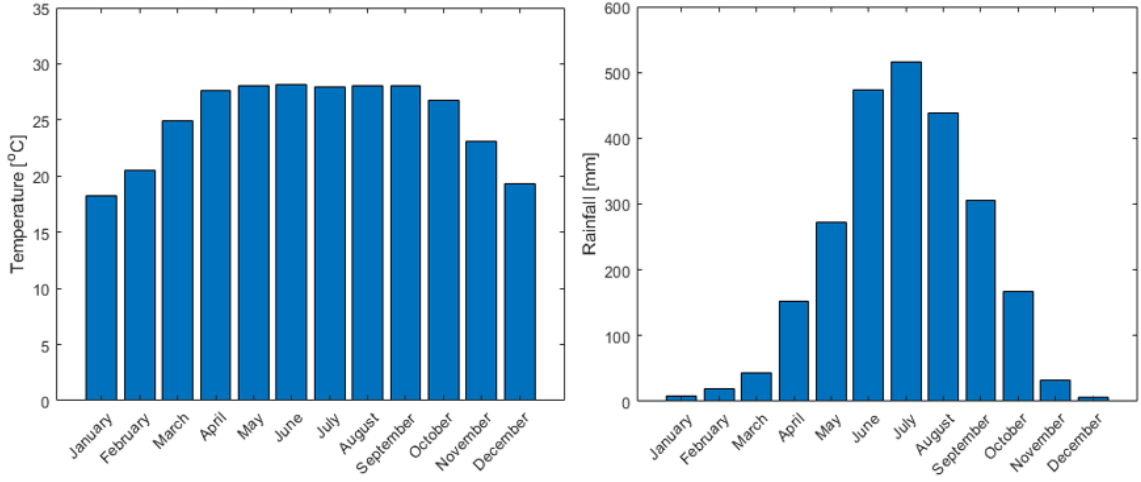

Fig. 2. Average (month-wise) temperature (left) and rainfall (right) of Bangladesh

The primary difference lies within the architecture of the memory cells. A memory cell of an LSTM network is much more intricate as opposed to an RNN module. A typical RNN module employs only the tanh function while an LSTM module employs several tanh and sigmoid functions. Figure 3 represents the standard RNN module. An LSTM module also employs a gates mechanism that allows information regulation within the memory cells. Besides, the gates also regulate the duration of information storage. The module constitutes 3 different types of gates; the input gate ($i_t$), the forget gate ($f_t$), and the output gate ($o_t$). Every memory cell also has a cell state that allows the flow of information and is interconnected to the gates. The gates regulate the amount of information that is fed to the cell state ($C_t$). At a particular time, t all of the 3 gates are fed with the following parameters; the new input ($x_t$) and the output or hidden state of the previous time step ($h_{t-1}$). The old cell state ($C_{t-1}$) is also incorporated within the memory cell. The forget gate ($f_t$) serves the purpose of determining the amount of the previous information or old hidden state ($h_{t-1}$) that is to be erased and how much of it is to be let through and stored in a cell state. The input gate ($i_t$) determines the amount of new input ($x_t$) that is to be let through and stored in a cell state. Finally, the output gate ($o_t$) determines the amount of information that is to be derived and filtered from the current cell state and used as a new hidden state ($h_t$). Figure 4 displays a basic LSTM memory cell.

The functions/operations according to [21] of the three gates are given below,

$$f_t = \sigma\left(W_f \times [h_{t-1}, x_t] + b_f\right) \quad (1)$$

The σ refers to the logistic sigmoid function and uses the new input ($x_t$) and previous hidden state ($h_{t-1}$) in order to determine what to let through. Here, $f_t$ gives a result that is in vector format. $W_f$ and $b_f$ are also sets of vectors called weights and bias which are learnable parameters and can be regulated accordingly for the forget gate.

$$\sigma(x) = \frac{1}{1+e^{-x}} \quad (2)$$

A typical sigmoid function is shown above and it gives resulting values ranging from 0 to 1.

$$i_t = \sigma\left(W_i \times [h_{t-1}, x_t] + b_i\right) \quad (3)$$

Here, $i_t$ gives a resulting vector that indicates the input gate. $W_i$ and $b_i$ are sets of vectors called weights and bias which are learnable parameters and can be regulated accordingly for the input gate.

$$\widehat{C}_t = \tanh \times \left(W_c \times [h_{t-1}, x_t] + b_c\right) \quad (4)$$

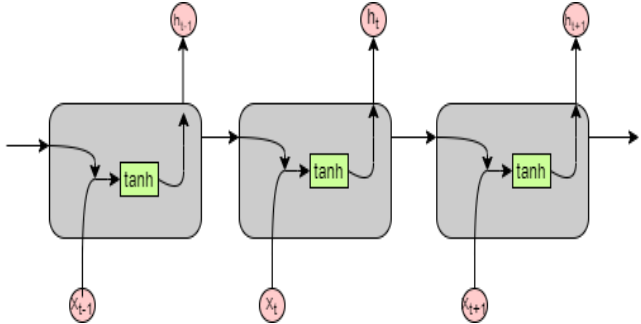

Fig. 3. Standard RNN module

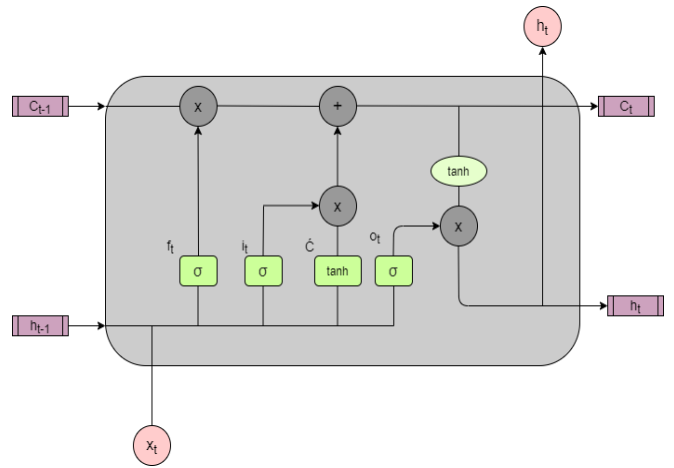

Fig. 4. Structure of the LSTM memory cell

$\hat{C}_t$ is known as the potential update vector whose output refers to amount of new information that is to be fed to the new cell state $C_t$. $W_c$ and $b_c$ are sets of vectors called weights and bias which are learnable parameters and can be regulated accordingly.

$$\tanh = \frac{e^x - e^{-x}}{e^x + e^{-x}} \quad (5)$$

In addition, tanh refers to the hyperbolic tangent function whose result is given by the equation above and gives output values ranging from -1 to 1.

$$C_t = f_t * C_{t-1} + i_t * \hat{C}_t \quad (6)$$

Equation for $C_t$ here gives the final output for the new cell state where '*' indicates element to element multiplication of the vectors or matrices.

$$o_t = \sigma(W_0 \times [h_{t-1}, x_t] + b_0) \quad (7)$$

"$o_t$" refers to the resulting vector of output gate which is a filtered version of the new cell state output ($C_t$) that is to be fed to the new hidden state ($h_t$). $W_o$ and $b_o$ are sets of vectors called weights and bias which are learnable parameters and can be regulated accordingly for the output gate. And $h_t$ is given by the equation below.

$$h_t = o_t * \tanh(C_t) \quad (8)$$

It is worth mentioning that it is due to the cell state's ($C_t$) linear interaction with the contents of the memory block that it is able to withhold information over a long period of time.

*C. Fundamental LSTM Blocks*

As evident from [22], basic LSTM is comprised of the memory cell and fully connected layers. Figure 5 displays the structure of a basic LSTM block.

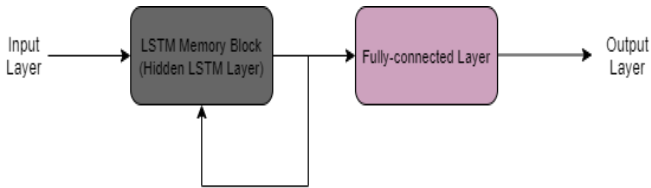

Fig. 5. Schematic of basic LSTM block

The overall architecture is comprised of a hidden layer of LSTM memory cells and a fully connected layer. This is shown in Figure 6. Successively, the layers are also connected to several input and output nodes. The output of the final LSTM layer depicts itself in vector format. The role of a fully connected layer is to ensure improved extraction and format for the output vector and also connect this vector to the final prediction.

The equations involved in the computation are given below.

$$(h_i, m_i) = \text{LSTM}\left(\begin{bmatrix} Input \\ h_{i-1} \end{bmatrix}, m, W\right)$$

$$\text{prediction} = \sigma(W^{fc} \times h_l + b^{fc}) \quad (9)$$

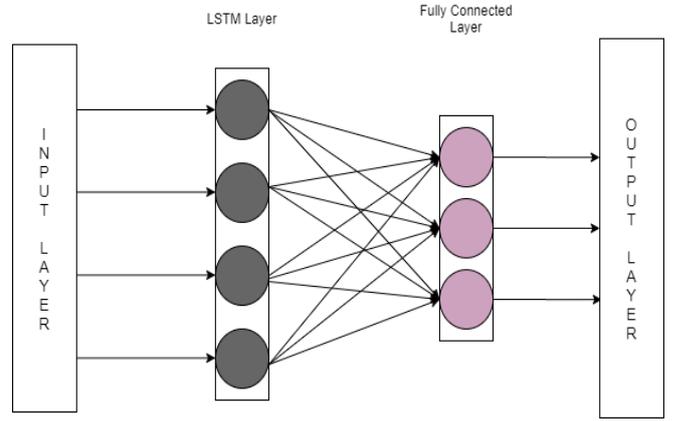

Fig. 6. Basic LSTM block architecture

In the equations provided above, $h_l$ is the hidden vector of final time-step LSTM, $W^{fc}$ represents the weight matrices within the fully connected layer and $b^{fc}$ represents the bias parameters.

### III. RESULTS AND DISCUSSION

Our implemented LSTM network is trained by employing Adam optimizer as the optimization algorithm using a learning rate of 0.001 and 100 hidden layers. To evaluate the model's performance, the mean squared error (MSE) is adopted as a loss function. Throughout our experiment, we used a workstation with the NVIDIA GeForce RTX 2080Ti GPU. The LSTM model was executed using PyTorch, running on Intel Xeon E5-2620, Core i5-2.4GHz processor, 16GB RAM.

The month-wise average temperature and rainfall are presented in Figure 2. It shows that the higher temperature and rainfall prevail during the mid of a year. Towards January and December, the rainfall is observed close to 0 mm. On the other hand, the month-wise average temperature varies less than 38°C around the year. We have used the weather data for 113 years (1901-2013) to train the LSTM model and 2 years (2014 and 2015) to test the forecasting performance of the model. The model showed significant accuracy in the case of forecasting the weather pattern. In the left part of Figure 7, we can observe the actual and predicted temperature for the years 2014 and 2015. It is evident that the model has learned the temperature pattern around a year which reflected on its prediction. The same goes for the forecasting of the rainfall which is showed in the right part of Figure 7.

TABLE I. LSTM MODEL PERFORMANCE

|  | Temperature prediction | Rainfall prediction |
|---|---|---|
| Mean error | -0.38 | -17.64 mm |
| Standard deviation of error | 0.63 | 145.13 mm |
| Absolute deviation of error | 0.53 ± 0.50 | 107.11 ± 97.05 mm |

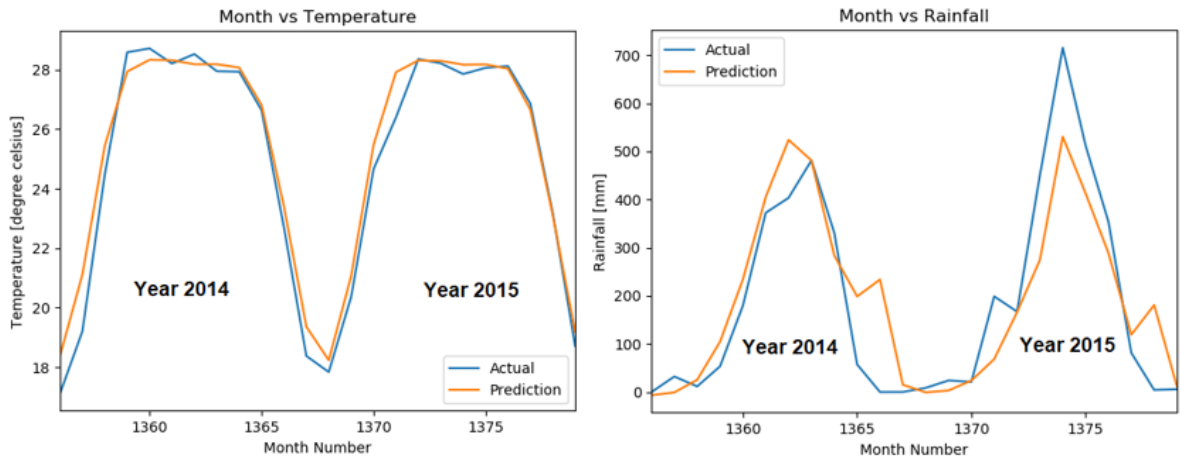

Fig. 7. Actual vs. predicted temperature (left) and rainfall (right) in years 2014 and 2015

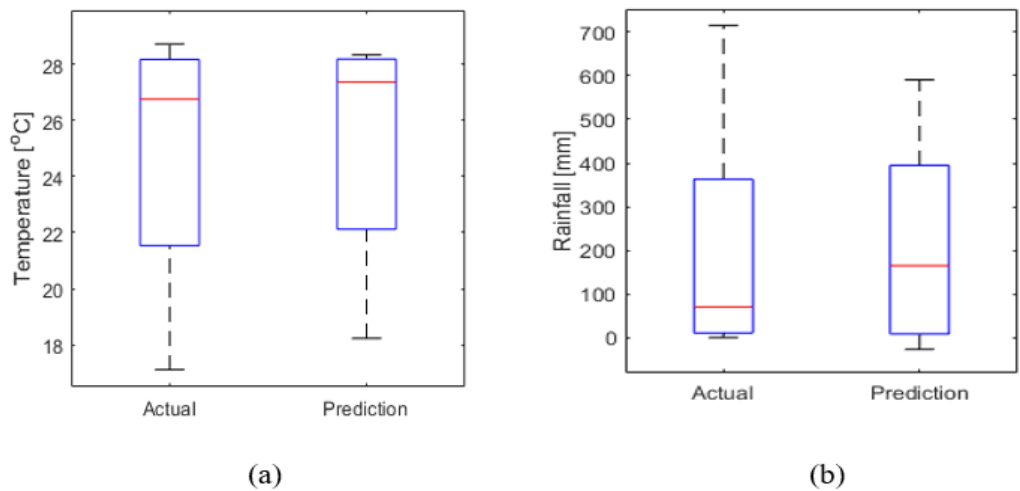

Fig. 8. Boxplot of actual and predicted (a) temperature (b) rainfall in the years 2014 and 2015

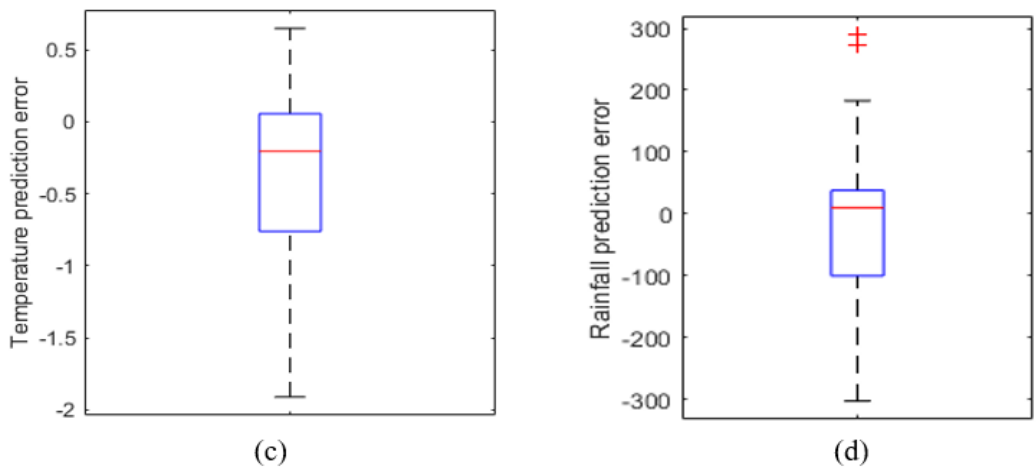

Fig. 9. Boxplot of actual and predicted (c) temperature prediction error (d) rainfall prediction error in the years 2014 and 2015

Figure 8 (a) and 8 (b) shows the actual and the predicted temperature and rainfall, respectively, for the year 2014 and 2015. In addition to that, Figure 9 (b) and 9 (c) show the prediction error after implementing the LSTM model. The model performance is depicted in Table 1. The mean error in the case of Temperature prediction implementing the LSTM model is very low (-0.38°C). Correspondingly, the mean prediction error for Rainfall is -17.64 mm which is low considering the wide range of actual rainfall (0.12 mm to 715.22 mm). The training losses for both the temperature and the rainfall prediction model are presented in Figures 10 and 11. The loss for the temperature prediction converges after 25 epochs. On the other hand, the loss for rainfall prediction converges after 40 epochs.

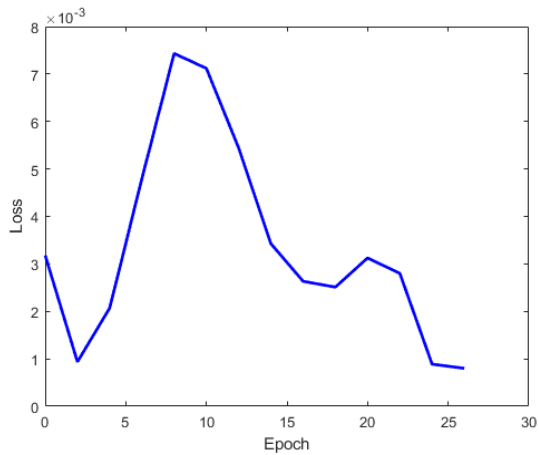

Fig. 10. Training loss vs. epoch for temperature prediction

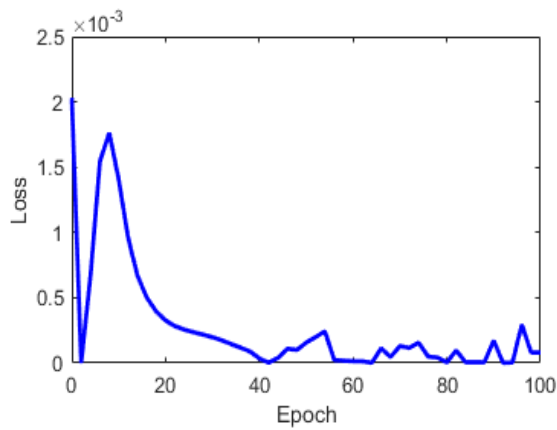

Fig. 11. Training loss vs. epoch for rainfall prediction

## IV. CONCLUSION

Accurate weather prediction is important in a country like Bangladesh as extreme weather conditions have led to the spread of diseases and in some situations, even death. If correctly predicted, preventive measures could be taken against such consequences. In this study, 115 years of month-wise average temperature and rainfall data of Bangladesh were used for the purpose of predicting it by the application of the LSTM model. For training the model, 113 years (1901-2013) of weather data was used and the remaining 2 years' (2014 and 2015) was utilized for testing the weather prediction performance of the trained model. A very low mean error of -0.38°C was found for the predicted temperatures and a considerably low mean prediction error of -17.64 mm was found for the rainfall. Additionally, the standard deviation error was also found to be considerably low as compared to the actual rainfall and temperature. Therefore, this model can be considered as a suitable model for the prediction of monthly rainfall and temperature of Bangladesh. However, in the future, deep learning models can be applied for this time series analysis for further error rate minimization and faster computation.

REFERENCES

[1] S. Curceac, C. Ternynck, T. B. M. J. Ouarda, F. Chebana, and S. D. Niang, "Short-term air temperature forecasting using Nonparametric Functional Data Analysis and SARMA models," *Environ. Model. Softw.*, vol. 111, pp. 394–408, 2019.

[2] K. Streit *et al.*, "Soil warming alters microbial substrate use in alpine soils," *Glob. Chang. Biol.*, vol. 20, no. 4, pp. 1327–1338, 2014.

[3] R. L. E. C. JL *et al.*, "GCTE-NEWS (2001) A meta-analysis of the response of soil respiration, net nitrogen mineralization, and aboveground plant growth to experimental ecosystem warming," *Oecologia*, vol. 126, pp. 543–562.

[4] K. Manideep and K. R. Sekar, "Rainfall prediction using different methods of Holt winters algorithm: A big data approach," *Int. J. Pure Appl. Math*, vol. 119, pp. 379–386, 2018.

[5] A. Graham and E. P. Mishra, "Time series analysis model to forecast rainfall for Allahabad region," *J. Pharmacogn. Phytochem.*, vol. 6, no. 5, pp. 1418–1421, 2017.

[6] S. Mehdizadeh, J. Behmanesh, and K. Khalili, "New approaches for estimation of monthly rainfall based on GEP-ARCH and ANN-ARCH hybrid models," *Water Resour. Manag.*, vol. 32, no. 2, pp. 527–545, 2018.

[7] Y. Dash, S. K. Mishra, and B. K. Panigrahi, "Rainfall prediction of a maritime state (Kerala), India using SLFN and ELM techniques," in *2017 International Conference on Intelligent Computing, Instrumentation and Control Technologies (ICICICT)*, 2017, pp. 1714–1718.

[8] T. Fischer and C. Krauss, "Deep learning with long short-term memory networks for financial market predictions," *Eur. J. Oper. Res.*, vol. 270, no. 2, pp. 654–669, 2018.

[9] Y. Zhu, X. Gao, W. Zhang, S. Liu, and Y. Zhang, "A bi-directional LSTM-CNN model with attention for aspect-level text classification," *Futur. Internet*, vol. 10, no. 12, p. 116, 2018.

[10] J. Lei, C. Liu, and D. Jiang, "Fault diagnosis of wind turbine based on Long Short-term memory networks," *Renew. energy*, vol. 133, pp. 422–432, 2019.

[11] J. Zhao, F. Deng, Y. Cai, and J. Chen, "Long short-term memory-Fully connected (LSTM-FC) neural network for PM2. 5 concentration prediction," *Chemosphere*, vol. 220, pp. 486–492, 2019.

[12] C. Hu, Q. Wu, H. Li, S. Jian, N. Li, and Z. Lou, "Deep learning with a long short-term memory networks approach for rainfall-runoff simulation," *Water*, vol. 10, no. 11, p. 1543, 2018.

[13] C. Li, Y. Zhang, and G. Zhao, "Deep Learning with Long Short-Term Memory Networks for Air Temperature Predictions," in *2019 International Conference on Artificial Intelligence and Advanced Manufacturing (AIAM)*, 2019, pp. 243–249.

[14] W. Liao, Z. YIN, R. Wang, and X. LEI, "RAINFALL-RUNOFF MODELLING BASED ON LONG SHORT-TERM MEMORY (LSTM)," 2019.

[15] D. Kumar, A. Singh, P. Samui, and R. K. Jha, "Forecasting monthly precipitation using sequential modelling," *Hydrol. Sci. J.*, vol. 64, no. 6, pp. 690–700, 2019.

[16] S. Poornima and M. Pushpalatha, "Prediction of rainfall using intensified lstm based recurrent neural network with weighted linear units," *Atmosphere (Basel).*, vol. 10, no. 11, p. 668, 2019.

[17] S. H. I. Xingjian, Z. Chen, H. Wang, D.-Y. Yeung, W.-K. Wong, and W. Woo, "Convolutional LSTM network: A machine learning approach for precipitation nowcasting," in *Advances in neural information processing systems*, 2015, pp. 802–810.

[18] F. Kratzert, D. Klotz, C. Brenner, K. Schulz, and M. Herrnegger, "Rainfall--runoff modelling using long short-term memory (LSTM) networks," *Hydrol. Earth Syst. Sci.*, vol. 22, no. 11, pp. 6005–6022, 2018.

[19] "No Title." https://www.kaggle.com/yakinrubaiat/bangladesh-weather-dataset.

[20] S. Hochreiter and J. Schmidhuber, "Long short-term memory," *Neural Comput.*, vol. 9, no. 8, pp. 1735–1780, 1997.

[21] C. Olah, "Understanding lstm networks, 2015," *URL http//colah. github. io/posts/2015-08-Understanding-LSTMs*, 2015.

[22] Q. Zhang, H. Wang, J. Dong, G. Zhong, and X. Sun, "Prediction of sea surface temperature using long short-term memory," *IEEE Geosci. Remote Sens. Lett.*, vol. 14, no. 10, pp. 1745–1749, 2017.